\documentclass[10pt,twocolumn,letterpaper]{article}

\usepackage{cvpr}              

\usepackage{graphicx}
\usepackage{amsmath}
\usepackage{amssymb}
\usepackage{booktabs}

\usepackage[pagebackref,breaklinks,colorlinks]{hyperref}

\usepackage[accsupp]{axessibility}  

\usepackage[capitalize]{cleveref}
\crefname{section}{Sec.}{Secs.}
\Crefname{section}{Section}{Sections}
\Crefname{table}{Table}{Tables}
\crefname{table}{Tab.}{Tabs.}

\def\cvprPaperID{79} 
\def\confName{CVPR}
\def\confYear{2023}

\begin{document}

\title{SB-VQA: A Stack-Based Video Quality Assessment Framework for Video Enhancement}

\author{Ding-Jiun Huang \\ {\small KKCompany Technologies, National Taiwan University} \\ {\tt\small starwars78910@gmail.com} 
\and Yu-Ting Kao \\ {\small KKCompany Technologies} \\ {\tt\small yvonnekao@kkcompany.com} 
\and Tieh-Hung Chuang  \\  {\small KKCompany Technologies} \\ {\tt\small ironheadchuang@kkcompany.com}
\and Ya-Chun Tsai \\ {\small KKCompany Technologies} \\ {\tt\small yachuntsai@kkcompany.com} 
\and Jing-Kai Lou \\ {\small KKCompany Technologies} \\ {\tt\small kaelou@kkcompany.com} 
\and Shuen‑Huei Guan \\ {\small KKCompany Technologies} \\ {\tt\small drakeguan@kkcompany.com} 
}







\maketitle

\begin{abstract}
In recent years, several video quality assessment (VQA) methods have been developed, achieving high performance. However, these methods were not specifically trained for enhanced videos, which limits their ability to predict video quality accurately based on human subjective perception. To address this issue, we propose a stack-based framework for VQA that outperforms existing state-of-the-art methods on VDPVE, a dataset consisting of enhanced videos.

In addition to proposing the VQA framework for enhanced videos, we also investigate its application on professionally generated content (PGC). To address copyright issues with premium content, we create the PGCVQ dataset, which consists of videos from YouTube. We evaluate our proposed approach and state-of-the-art methods on PGCVQ, and provide new insights on the results. Our experiments demonstrate that existing VQA algorithms can be applied to PGC videos, and we find that VQA performance for PGC videos can be improved by considering the plot of a play, which highlights the importance of video semantic understanding.

\end{abstract}

\section{Introduction}
\label{sec:intro}

Video quality assessment (VQA) aims to evaluate the quality of videos conforming to human subjective perception, with subjective feedback, such as mean opinion score (MOS), as measurement. In recent years, many VQA methods have been proposed, and have achieved high performance on various benchmarks \cite{zhang2018blind,sinno2018large,hosu2017konstanz,nuutinen2016cvd2014}. 

Video enhancement techniques, such as video super resolution (VSR) and video restoration, aim to improve the quality of videos by generating high-resolution videos or removing artifacts while objective metrics, such as PSNR and SSIM \cite{wang2004image}, are commonly used to evaluate their performance. Since these metrics for video enhancement are pixel-based and cannot really reflect subjective perception, an accurate VQA method would be beneficial to video enhancement research. 

Previous research on VQA for video enhancement is limited. To address this gap, we propose the Stack-Based Video Quality Assessment network (SB-VQA). The SB-VQA consists of the following components: feature extractors using FANet \cite{wu2022fast}, patch-weighted convolution blocks, and a final regression block. Firstly,  a given video is sampled with grid mini-patch sampling (GMS) \cite{wu2022fast} for computational efficiency. The input is then processed by three stacked feature extractors that perform self-attention independently. Each extractor is followed by a dual-branch
convolution block for better quality prediction. Finally, a regression block generates the final prediction score by considering the outputs from each branch. SB-VQA is highly scalable, allowing for an increase or decrease in the number of branches depending on the available computational resources. 

To evaluate the performance of SB-VQA, we benchmarked our method against VQA datasets, including VDPVE \cite{gao2023vdpve}, a recently proposed video enhancement dataset. VDPVE consists of $1,211$ videos processed by a wide range of enhancement methods such as color correction, brightness adjustment, contrast enhancement, deblurring, and deshaking including recent state-of-the-art methods like BasicVSR++ \cite{chan2022basicvsr++}. The results show that SB-VQA outperforms existing VQA methods when fine-tuned on both VDPVE and other datasets.

Since most video datasets, including VDPVE, contain only user generated content (UGC), we wonder whether existing state-of-the-art VQA methods can also have good performance on professionally generated content (PGC). PGC videos, mostly produced in the film or television industry, usually has much better quality compared with UGC videos. Although PGC videos hardly contain distortions, many applications such as VSR or video restoration on old films require an accurate VQA approach on PGC videos for benchmark. Therefore, we construct a PGC dataset for video quality assessment (PGCVQ). PGCVQ contains official movie trailers from YouTube with the Creative Commons license. The PGC videos are encoded with different bitrates, which is common in the streaming industry. We also retrieve the heatmap information from YouTube videos \cite{google}, which roughly represents viewers’ preference for content. Our experiments on PGCVQ have two main goals: (1) to verify the performance of existing state-of-the-art VQA on PGC videos, and (2) to analyze the impact of picture quality (resolution, degree of distortion) and human interests. For the first goal, $1,200$ videos are randomly sampled from PGCVQ to make predictions using our proposed SB-VQA. For the second goal, we split a trailer into numerous segments and predict quality scores on each segment. We then analyze the results using heatmap to explore the relationship between VQA and video content.

To sum up, our work has following contributions: (1) We propose SB-VQA, a scalable stack-based VQA framework that outperforms existing state-of-the-art algorithms on benchmarks including VDPVE. (2) We construct PGCVQ, a VQA dataset of professionally generated content. It can benefit VQA, VSR, or video restoration research in the future. Moreover, we verify the feasibility of applying existing VQA methods on PGC videos. (3) We analyze the VQA task with PGC videos and video heatmaps.  We conclude that the content of a video does affect human subjective assessment of quality.

\section{Related Works}
\label{sec:related}
\noindent\textbf{No-reference image and video quality assessment}
No-reference quality assessment investigates the “quality” of individual images or videos by themselves without any reference directly. At beginning, based-on the assumptions that the natural scene statistics (NSS) extracted from natural images are highly regular, several no-reference image quality assessment (IQA) methods have been developed such as NIQE \cite{mittal2012making} and BRISQUE \cite{mittal2012no}. Deep learning-based no-reference IQA methods further improve the performance in terms of real-world distortions. MANIQA \cite{yang2022maniqa} outperforms state-of-the-art methods on several standard datasets by a large margin with integrating modules of ViT \cite{dosovitskiy2020image} and Swin Transformer \cite{liu2021swin}. For no-reference VQA, a naive method is to compute the quality of each frame via no-reference IQA methods, and then pool them into the video quality score. A comparative study of various temporal pooling strategies on popular no-reference IQA methods can refer to \cite{tu2020comparative}. Compared to no-reference IQA, no-reference VQA further requires the characterization of quality-aware temporal distortion, which amplifies the difficulty of quality assessment. Some research combines the findings on the human visual system, for example, TPQI is proposed to measure the temporal distortion by describing the graphical morphology of the representation \cite{liao2022exploring}. Recently, due to the nature of subjection of quality assessment, researchers attempt to explain a single score of no-reference VQA into multiple perspectives \cite{wu2022disentangling,wang2021rich}. For example, DOVER further separates the quality of user generated content (UGC) videos based on the technical and aesthetic perspectives, where the technical perspective, measuring the perception of distortions, and the aesthetic perspective, which relates to preference and recommendation on contents \cite{wu2022disentangling}.  

\noindent\textbf{Fixed-feature-based video quality assessment}
In the realm of video quality assessment (VQA), deep neural networks (DNNs) have shown potential for achieving high accuracy. However, the computational cost of using DNNs on high resolution videos can be prohibitively high. To address this challenge, many VQA methods opt to train a feature regression network using fixed deep features \cite{li2021unified, sun2022deep}. For instance, SimpleVQA \cite{sun2022deep} leverages spatial features from a ResNet50 \cite{he2016deep} backbone trained on ImageNet \cite{deng2009imagenet}, as well as motion features from a pre-trained SlowFastR50 \cite{feichtenhofer2019slowfast} model trained on the Kinetics 400 dataset \cite{kay2017kinetics}, and then regresses the features into quality scores. Other VQA methods \cite{ying2020patches,li2022blindly} use the feature extractors pre-trained with IQA datasets \cite{ying2020patches,ciancio2010no,ghadiyaram2015massive,hosu2020koniq,fang2020perceptual}. By transferring knowledge from IQA databases with authentic distortions, these methods are able to generate quality-aware spatial feature representations without relying on pre-trained DNNs \cite{li2022blindly}. However, the computational cost of these approaches remains high for high resolution videos, and their accuracy may be limited by the use of fixed features that are not optimized for quality-related information extraction. 

\begin{figure*}[hbt!]
  \centering
   \includegraphics[width=0.8\linewidth]{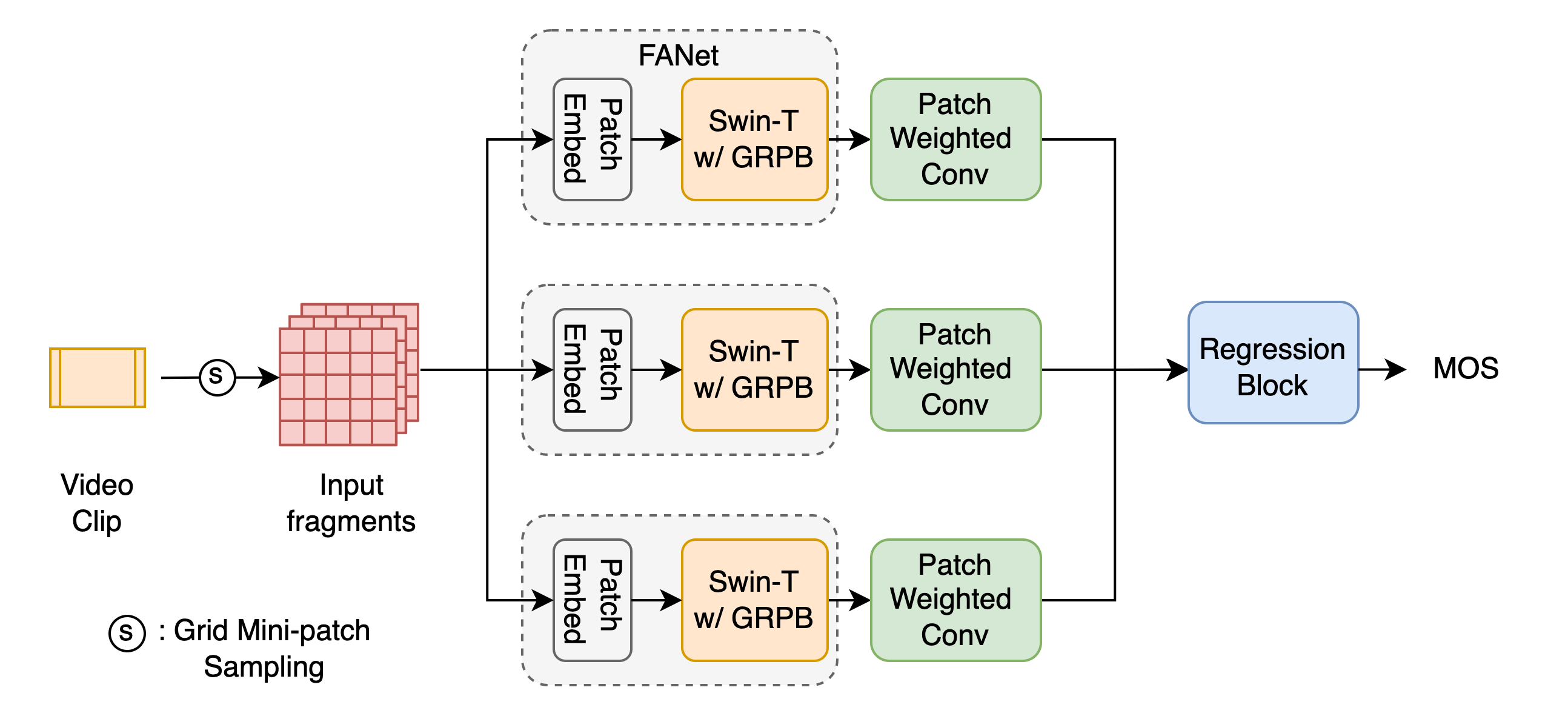}

   \caption{Overall framework of SB-VQA}
   \label{fig:framework}
\end{figure*}

\noindent\textbf{End-to-end deep video quality assessment}
Compared with fixed feature extraction, end-to-end training enables learning video-quality-related representations that better represent quality information, which is believed to improve the accuracy. Several DNNs are designed for  jointly end-to-end learning of features and regression based on the raw input data \cite{zhang2018blind,bosse2017deep}. However, these methods suit only on low resolution videos; applying high resolution videos directly to the existing methods leads to memory shortage problems on GPU. Fast-VQA is therefore designed, which samples mini-patches in uniform grids at raw resolution to improve the computational efficiency while maintaining the global quality \cite{wu2022fast}. Though the efficiency and the results are impressive over existing datasets, how it would perform on the VQA for video enhancement and video compression in high resolution videos remains unknown.

\noindent\textbf{Quality assessment of video compression for PGC}
Most existing works focus on VQA for UGC videos, and only limited works address VQA for PGC videos. One possible reason may lie on the copyright issue. To investigate how VQA performs on PGC, a dataset consisting of legal PGC videos is required for benchmarking. To tackle this problem, in this work, we build a PGC dataset based on the official trailers released on Youtube as they are open to the public.

\section{Approach}
\label{sec:approach}

In this section, we present the overall pipeline of our proposed Stack-Based Video Quality Assessment (SB-VQA) method for evaluating the quality of enhanced videos. To address the challenge of bias introduced by a wide range of video enhancements, we develop a novel stack-based framework for quality prediction, with a two-stage training strategy: a transformer-based block to predict the quality directly from a sequence of frames, and then a regression block to map the outputs of feature extractors to a final quality score. In the transformer-based block, we incorporate Grid Mini-Patch Sampling (GMS) and Fragment Attention Network (FANet) of Fast-VQA \cite{wu2022fast} to extract both spatial and temporal information from videos. Then, inspired by MANIQA \cite{yang2022maniqa}, we further incorporate a dual branch convolution block into the FANet. The details are introduced in the following subsections.

\subsection{Overall Framework}
\label{sec:overall_framework}

\cref{fig:framework} shows the overall framework of SB-VQA. To mitigate the negative effects of bias caused by diverse enhancements in videos from the training set \cite{LeBlanc1996CombiningEI}, we propose a stack-based framework for the quality assessment. Inputs are first sampled from videos using GMS, and then fed into feature extractors which are fine-tuned with different data splits and hyperparameters. A following regression block maps the outputs of feature extractors to a final quality score.
 
We train the framework consisting of $K$ different feature extractor models in two stages. First, the training set will be divided into $J$ folds. For each feature extractor model $k$, $J$ copies of $k$ will be trained. The $j_{th}$ copy is trained with \{$i_{th}$ fold $\mid$ $i\leq J$ and $i\neq j$ \}, and will predict on the $j_{th}$ fold. The predictions of all the folds can be served as $k$’s prediction on the training set. In the second stage, the $K$ predictions, as well as their labels, will be taken as meta-data to train the regression meta-model. In the testing phase, predictions from all $k_{th}$ model’s copies are averaged, and the $K$ averaged predictions are fed into the regression block for final quality score.

\subsection{Grid Mini-patch Sampling (GMS)}

In order to improve computational efficiency while preserving the quality of original video, we use the GMS method proposed in Fast-VQA to sample input videos to “fragments”. First, each frame of a video is divided into uniform grids. Then, a mini-patch is randomly selected from every grid. Finally, the mini-patches are spliced into a “fragment”, serving as the input of the following attention layers. Note that the sampled mini-patch with original resolution is aligned in all the frames, so temporal as well as spatial information can be well preserved. 

\begin{figure}[hbt!]
  \centering
   \includegraphics[width=1\linewidth]{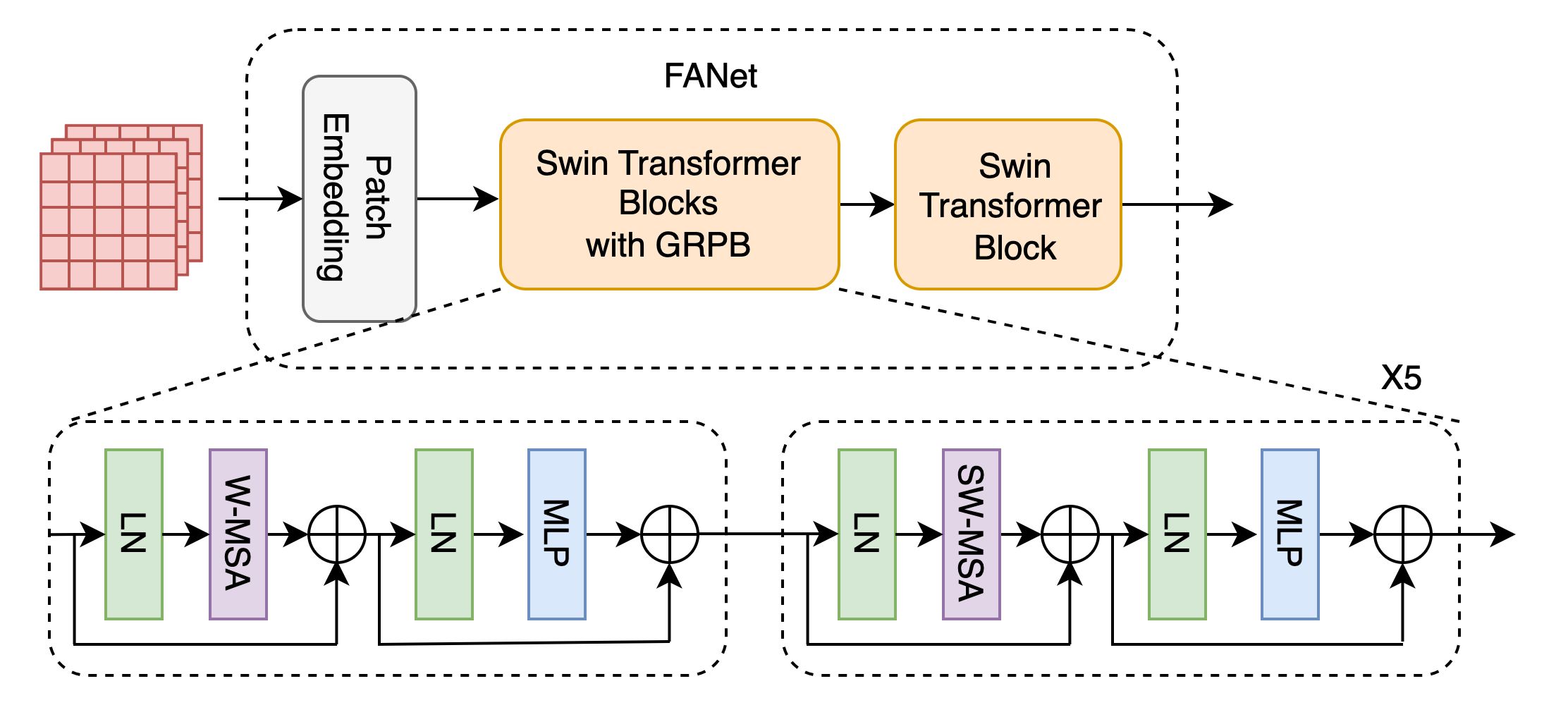}

   \caption{Feature Extractor of SB-VQA}
   \label{fig:feature_extractor}
\end{figure}

\subsection{Feature Extractor}

As shown in \cref{fig:feature_extractor}, the input fragment, spliced by the sampled mini-patches, is fed into FANet, which includes Swin Transformer blocks with Gated Relative Position Biases (GRPB) \cite{wu2022fast} to capture both local and global quality for feature extraction. Specifically, Gated Relative Position Biases (GRPB) is adopted in Swin Transformer Tiny (Swin-T) \cite{liu2022video} as the to obtain a learnable position bias table for intra-patch and cross-patch attention respectively because intra-patch attention pairs have much smaller actual distances than cross-patch attention pairs.

\subsection{Dual Branch for Patch-weighted Prediction}
\begin{figure}[hbt!]
  \centering
   \includegraphics[width=1\linewidth]{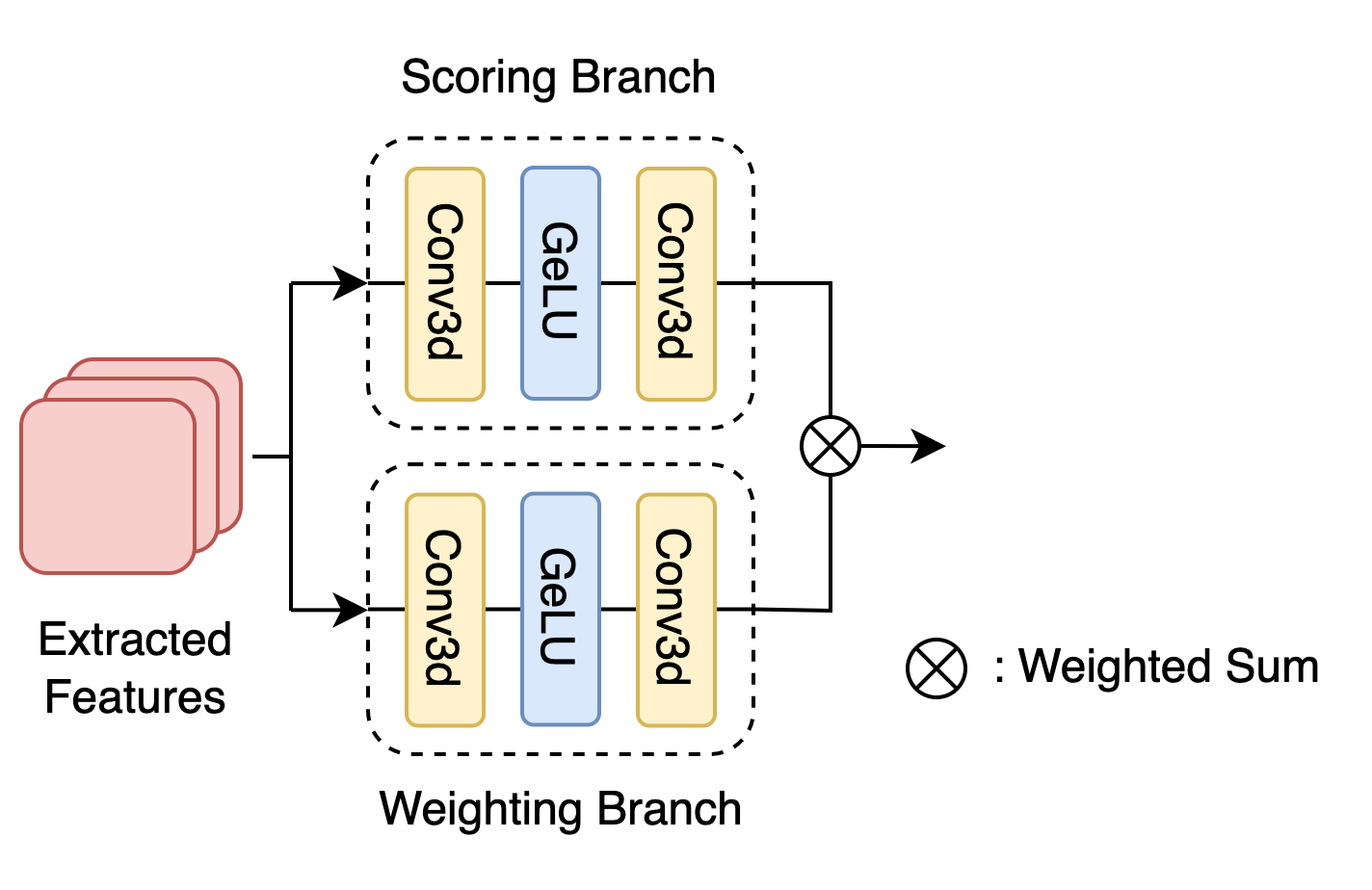}

   \caption{Dual branch structure for computing each patch's score and weight}
   \label{fig:dual_branch}
\end{figure}

When assessing the quality of videos, people tend to focus on certain regions of the videos, e.g., the center or moving objects in the video. Moreover, a patch with bad quality would have greater influence on quality score than a patch with good quality does. To address the issue that different regions on the resulting feature map have varying impacts on quality score, we integrate the dual-branch structure \cite{yang2022maniqa} into Intra-Patch Non-Linear Regression \cite{wu2022fast} to predict the weighted score from each patch, as shown in \cref{fig:dual_branch}. The structure consists of a scoring and a weighting branch, in which the scoring branch predicts each patch’s quality and the weighting branch estimates the importance of each patch. The weighted sum of the two branches leads to the quality score. Each branch consists of 3D convolution and GeLU layers. Given a feature F, the two branches output score and weight features respectively. Then, a quality score is calculated from the multiplication of every patch’s weight and score. We believe the dual branch structure of patch-weighted quality prediction is closer to human perceptions of videos.

\section{Experiment}
\label{sec:experiment}

\begin{table*}[hbt!]
  \centering
  \begin{tabular}{c r r c}
    \toprule
    Dataset & No. of Video & Video Duration (sec) & Video Resolution\\
    \midrule
    KoNViD-1k & 1,200 & 8 & 540p \\
    LIVE-VQC & 585 & 10 & 240p--1080p \\
    LSVQ & 39,075 & 5--12 & 99p--4k \\
    VDPVE & 1,211 & 8--10 & 720p, 1080p \\
    \bottomrule
  \end{tabular}
  \caption{Descriptive statistics for the included datasets.}
  \label{tab:summary}
\end{table*}

\begin{table*}[hbt!]
	\centering
	\begin{tabular}{c||c c |c c || c c}
		\hline\hline
		\ &\multicolumn{2}{|c}{KoNViD-1k}
        &\multicolumn{2}{|c||}{LIVE-VQC}
        &\multicolumn{2}{|c}{VDPVE$_{train}$}\\
		\cline{2-7}
		 Method &  SRCC & PLCC  & SRCC & PLCC & SRCC & PLCC\\
		\hline\hline
		TLVQM \cite{Korhonen2019TwoLevelAF} & 0.773 & 0.768 & 0.799 & 0.803 & - & -\\
		\hline
		VIDEVAL \cite{Tu2020UGCVQABB} & 0.783 & 0.780 & 0.752 & 0.751 & - & -\\
		\hline
		PVQ$_{w/patch}$ \cite{Ying2020PatchVQU} & 0.791 & 0.786 & 0.827 & 0.837 & - & -\\
		\hline
		VSFA \cite{Li2019Quality} & 0.773 & 0.775 & 0.773 & 0.795 & - & -\\
		\hline
		Fast-VQA \cite{wu2022fast} & 0.891 & 0.892 & 0.849 & 0.865 & 0.820 & 0.829\\
            \hline
            DOVER \cite{wu2022disentangling} & \textbf{0.909} & \textbf{0.906} & 0.860 & 0.875 & 0.840 & 0.841\\
		\hline
            SB-VQA (ours)  & 0.895 & 0.900 & \textbf{0.883} & \textbf{0.891} & \textbf{0.862} & \textbf{0.857}\\
		\hline\hline
		
	\end{tabular}
	\caption{Results of Fine-tuning LSVQ-trained models on different VQA datasets}
	\label{tab:finetune}
\end{table*}

\subsection{Dataset}

VDPVE \cite{gao2023vdpve} is a recently proposed VQA dataset, which contains videos processed by a variety of enhancement methods. The dataset contains $1,211$ videos in total, consisting of $600$ videos with color, brightness, and contrast enhancements, $310$ deblurred videos, $301$ deshaked videos, and MOS of each video. The videos are obtained from original videos selected from several datasets \cite{ghadiyaram2017capture,berns2019v3c1,hosu2017konstanz,sinno2018large,wang2019youtube,madhusudana2021subjective,mackin2018study,liu2013bundled} with $8$ enhancement methods, $5$ deblurring methods, and $7$ deshaking methods. VDPVE is also the dataset of NTIRE 2023 Quality Assessment of Video Enhancement Challenge, and as participant of the competition, we only have an average score of Pearson’s linear correlation coefficient (PLCC) and Spearman rank correlation coefficient (SRCC) on the test set. Therefore, here the average score is compared on VDPVE. We also conduct the experiments on LIVE-VQC \cite{sinno2018large}, KoNViD-1k \cite{hosu2017konstanz} and LSVQ \cite{zhang2018blind}. \cref{tab:summary} shows the summary of the datasets that are used in our experiment.

\subsection{Implementation Detail}
All of our experiments are implemented with PyTorch 1.13.1 and ran on Amazon EC2 with one NVIDIA A10G GPU. When applying Grid Mini-patch Sampling, we divide each video frame into $7 \times 7$ grids, and the size of a patch sampled from a grid is set to $32$. For FANet, a window size of $(8, 7, 7)$ is adopted. An AdamW optimizer is adopted, and the learning rate is $0.001$. In each branch of the patch-weighted convolution block, the input and output channels of the two 3D convolution layers are $(768, 128)$ and $(128,1)$ respectively. Considering the diverse data distribution of the VDPVE dataset, we choose the Extreme Gradient Boosting \cite{Chen2016XGBoostAS} for the final score regression block. 

Our stack-based framework consists of three branches with the introduced feature extractor and patch-weighted convolution block but different hyperparameters. Moreover, we replace FANet in one of the three branches with its variant from Faster-VQA \cite{Wu2022NeighbourhoodRS} to improve the performance in NTIRE 2023 VQA Challenge.

\subsection{Training \& Evaluation}

The framework is trained and fine-tuned in a two-stage process detailed in Sec. \ref{sec:overall_framework}. We first train the framework with LSVQ$_{train}$ dataset. Then, the VDPVE training set is split into several folds for fine-tuning. 

To evaluate the performance of our proposed framework, we compare it to other VQA approaches on the VDPVE test set. Additionally, we assess the generalization ability of our framework on other in-the-wild VQA datasets, such as KoNViD-1k and LIVE-VQC. We measure the performance using PLCC and SRCC metrics.

\subsection{Performance}
\label{sec:performance}
\noindent\textbf{VDVPE test set}
In \cref{tab:vdpve_test}, we compare the performance of several state-of-the-art VQA methods on the VDPVE test set, and all the methods are trained on the VDPVE training set. We use PLCC and SRCC to measure their performances. The main score is defined as $(SRCC + PLCC) / 2$, which determines the final rank in the NTIRE 2023 VQA Challenge. In the table, we can see that our approach has the best prediction performance on the VDPVE test set. Note that the VDPVE test set is not publicly disclosed, and participants of the challenge only get the main score of their final submission. We use the results from \cite{gao2023vdpve} for comparison.

\begin{table}[hbt!]
	\centering
	\begin{tabular}{c||c c | c}
		\hline\hline
            & \multicolumn{3}{|c}{VDPVE$_{test}$} \\
		\cline{2-4}
		  Method & SRCC & PLCC & Main Score\\
		\hline\hline
		VIDEVAL \cite{Tu2020UGCVQABB} & 0.5005 & 0.4724 & 0.4865\\
            \hline
            RAPIQUE \cite{Tu2021RAPIQUERA} & 0.5434 & 0.5393 & 0.5414\\
            \hline
		TLVQM \cite{Korhonen2019TwoLevelAF} & 0.5474 & 0.5509 & 0.5492\\
		\hline
		V-BLIINDS \cite{Saad2014BlindPO} & 0.5652 & 0.5503 & 0.5578\\
            \hline
            VSFA \cite{Li2019Quality} & 0.5871 & 0.5424 & 0.5648\\
		\hline
		BVQA \cite{li2022blindly} & 0.6995 & 0.6674 & 0.6835\\
		\hline
		Fast-VQA \cite{wu2022fast} & 0.7350 & 0.7310 & 0.7330\\
		\hline
            SB-VQA (ours) & - & - & \textbf{0.7635}\\
		\hline\hline
	\end{tabular}
	\caption{The official prediction performance of VQA approaches on the NTIRE 2023 VQA Challenge test set.}
	\label{tab:vdpve_test}
\end{table}

\noindent\textbf{Fine-tuning results on other datasets}
\cref{tab:finetune} shows the performances of VQA approaches trained on LSVQ and fine-tuned on different datasets. Inspired by the experiments from \cite{wu2022fast}, we divide each dataset into random splits for 10 times, and calculate the average performance on test split. Our approach outperforms most others in all datasets. Two state-of-the-art methods, Fast-VQA \cite{wu2022fast} and DOVER \cite{wu2022disentangling}, are also included in the comparison. \cref{tab:finetune} shows that our approach achieves higher accuracy on VDPVE than the existing state-of-the-art. It implies that our approach has better performance when fine-tuned on enhanced videos.

\noindent\textbf{Cross-dataset test set}
In \cref{tab:cross}, we show the results of cross-validation experiments, where models are trained on LSVQ and tested on different datasets. While our approach shows comparable performance with other approaches, it does not outperform the existing state-of-the-art. The possible reason is that the regression block in our framework causes an overfitting on the training/fine-tuning dataset. Therefore, our approach performs better when fine-tuned on a portion of the target dataset. 

\begin{table}[hbt!]
	\centering
	\begin{tabular}{c||c c |c c}
		\hline\hline
		\ &\multicolumn{2}{|c}{KoNViD-1k}
        &\multicolumn{2}{|c}{LIVE-VQC}\\
		\cline{2-5}
		 Method &  SRCC & PLCC  & SRCC & PLCC \\
		\hline\hline
		TLVQM \cite{Korhonen2019TwoLevelAF} & 0.732 & 0.724 & 0.670 & 0.691\\
            \hline
            VIDEVAL \cite{Tu2020UGCVQABB} & 0.751 & 0.741 & 0.630 & 0.640\\
            \hline 
		PVQ$_{w/patch}$ \cite{Ying2020PatchVQU} &  0.791 & 0.795 & 0.770 & 0.807\\
            \hline
            VSFA \cite{Li2019Quality} & 0.784 & 0.794 & 0.734 & 0.772\\
		\hline
		Fast-VQA \cite{wu2022fast} &  0.859 & 0.855 & 0.823 & 0.844\\
            \hline
            DOVER \cite{wu2022disentangling} & \textbf{0.884} & \textbf{0.883} & \textbf{0.832} & \textbf{0.855}\\
		\hline
            SB-VQA (ours) & 0.841 & 0.838 & 0.821 & 0.848\\
		\hline\hline
		
	\end{tabular}
	\caption{Results of cross-validation of LSVQ-trained models on different VQA datasets}
	\label{tab:cross}
\end{table}

\subsection{Ablation Study}

\begin{table}[hbt!]
	\centering
	\begin{tabular}{c||c c }
		\hline\hline
		\ &\multicolumn{2}{|c}{VDPVE$_{train}$}\\
		\cline{2-3}
		 Model Variants &  SRCC & PLCC \\
		\hline\hline
		FANet $w/$ IP-NLR &  0.820 & 0.829 \\

            FANet $w/$ Patch-weighted Block & 0.838 & 0.830 \\
		\hline
            SB-VQA $w/ 2 \ branches$ & 0.848 & 0.840 \\

            SB-VQA $w/ 3 \ branches$ & 0.862 & 0.857 \\
		\hline\hline
		
	\end{tabular}
	\caption{Ablation study on patch-weighted convolution block and stack-based strategy.}
	\label{tab:ablation}
\end{table}

In this section, we provide ablation experiments to prove the effectiveness of patch-weighted convolution block and the stack-based strategy of our framework. The model variants in \cref{tab:ablation} are first trained with LSVQ, then fine-tuned on VDPVE$_{train}$ with the method mentioned in Sec. \ref{sec:performance}.

In the first two rows of \cref{tab:ablation}, we compare the proposed patch-weighted convolution block with Intra-Patch Non-Linear Regression (IP-NLR) adopted in Fast-VQA. It is shown that our proposed module achieves higher performance, suggesting that the patch-weighted block leads to better prediction conforming human perception. 

We also analyze the effects of our stack-based strategy. We compare SB-VQA with different number of stacked branches in \cref{tab:ablation}. It is shown that the stack-based strategy leads to significant improvements, compared with FANet $w/$ patch-weighted block, which is actually SB-VQA with only one branch. It implies that the stack-based strategy is helpful in reducing the bias introduced by the diverse enhancement methods. 

\section{Professionally Generated Content Analysis}

In this section, we focus on the video quality assessment for professionally generated content (PGC) videos (PGC VQA). While PGC videos typically have fewer distortions than user-generated content (UGC) videos, accurate VQA is still essential for a variety of applications, including old film restoration and video super resolution (VSR). An accurate VQA method can provide human subjective feedback on VSR, leading to more visually satisfying results. Because applications such as VSR on PGC videos have great industrial value and few previous works focus on VQA for PGC videos, we want to develop a solution of VQA on PGC videos with high performance.

To evaluate the effectiveness of our proposed VQA framework on PGC videos, we construct a PGC videos dataset, PGCVQ, of movie trailers from YouTube. Then, we fine-tune our proposed SB-VQA on VDPVE dataset, and test on PGCVQ. Our aim is to verify whether the state-of-the-art VQA method trained on UGC data can also perform well on PGC data. Moreover, we analyze the VQA task with different perspectives: we want to figure out whether the quality of picture (resolution, degree of distortion) or video content is more influential to human observation. To our knowledge, this is the first work to discuss VQA methods on PGC videos.

\subsection{PGC Dataset Construction}

\begin{figure}[hbt!]
  \centering
   \includegraphics[width=1\linewidth]{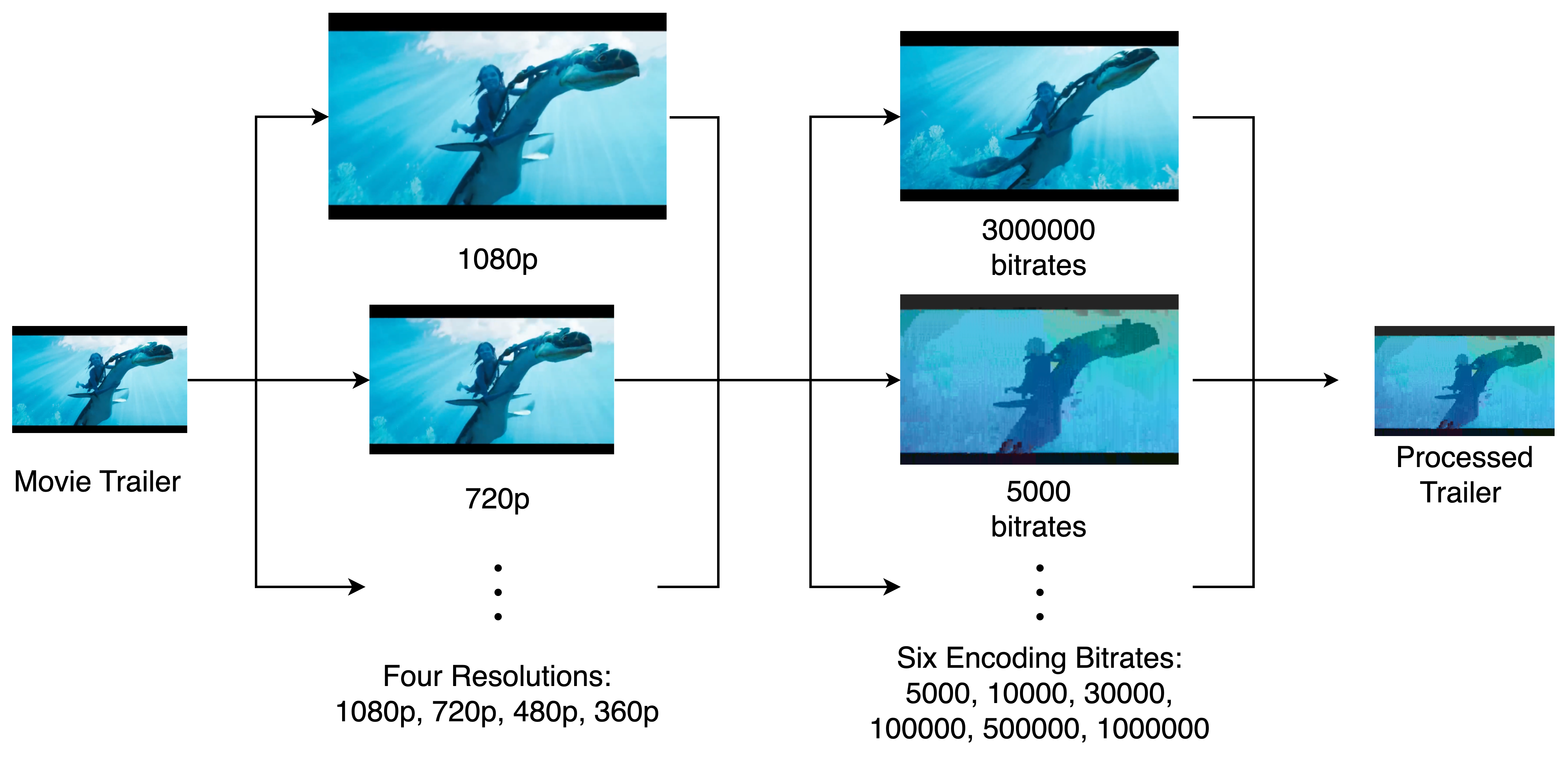}

   \caption{The pipeline of processing movie trailers from YouTube with different encoding bitrates and resolutions. \cite{Avatar2022}}
   \label{fig:pgc_pipeline}
\end{figure}

\subsubsection{Original videos}
\cref{fig:pgc_pipeline} shows the process of constructing the dataset. We first select official movie trailers released by movie studios on YouTube with the Creative Commons license. For each movie trailer, we obtain versions with different resolutions directly from YouTube. 

\subsubsection{Diversity in quality}
We transcoded the movie trailers of different content and resolutions with six bitrates. It’s well-known that higher encoding bitrates preserve more details and lead to better perceptual quality under given encoder type, video content, frame size and frame rate \cite{zhai2008cross}. Video compression with different encoding bitrates is common for PGC videos in the streaming industry \cite{rippel2019learned}, so our works have contributions to practical applications. 

\begin{figure*}[hbt!]
  \centering
  \begin{subfigure}{0.42\linewidth}
    \includegraphics[width=1\textwidth]{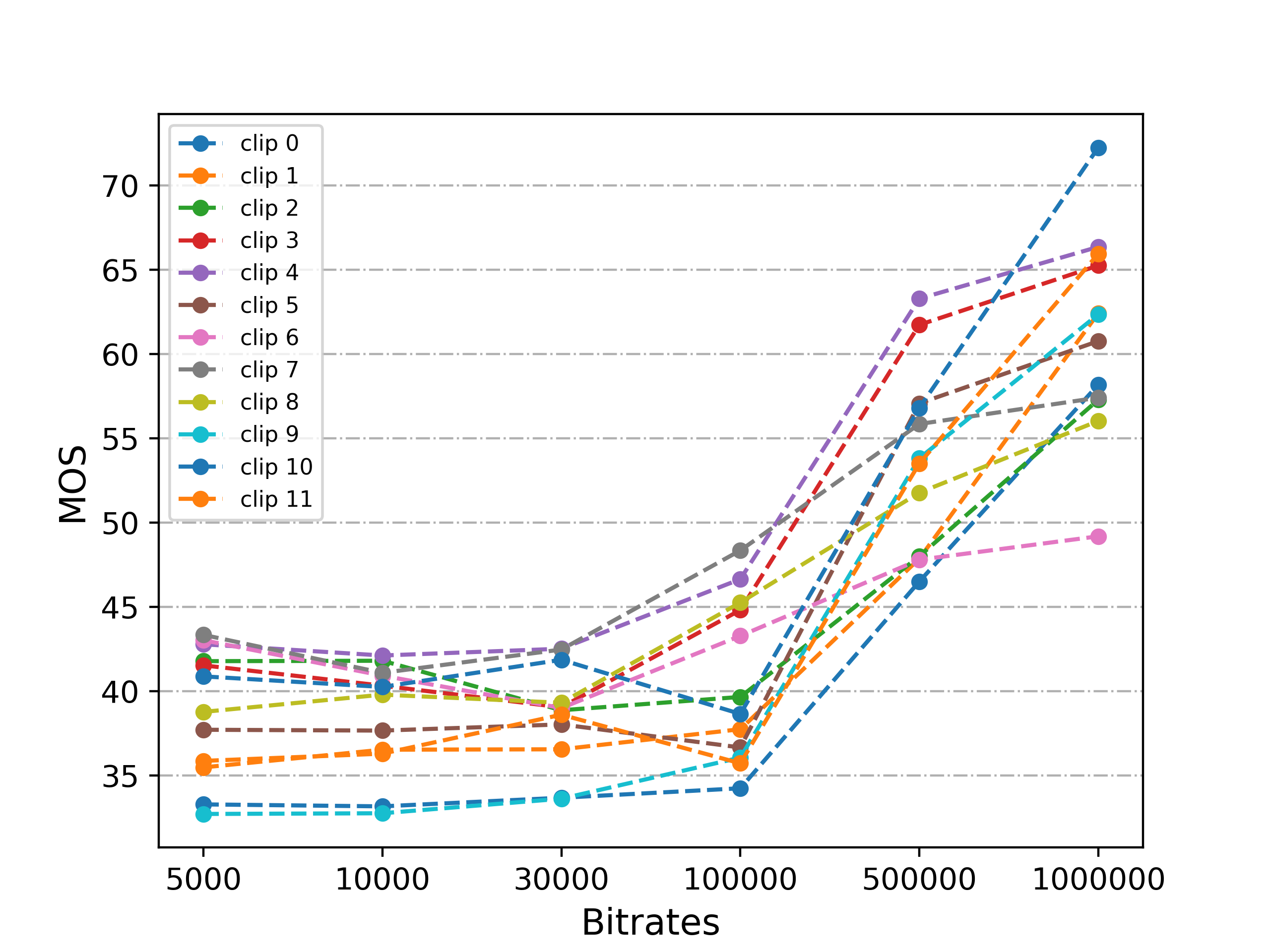}
    \caption{Predicted MOS of 1080p videos}
    \label{fig:1080p}
  \end{subfigure}
  \begin{subfigure}{0.42\linewidth}
    \includegraphics[width=1\textwidth]{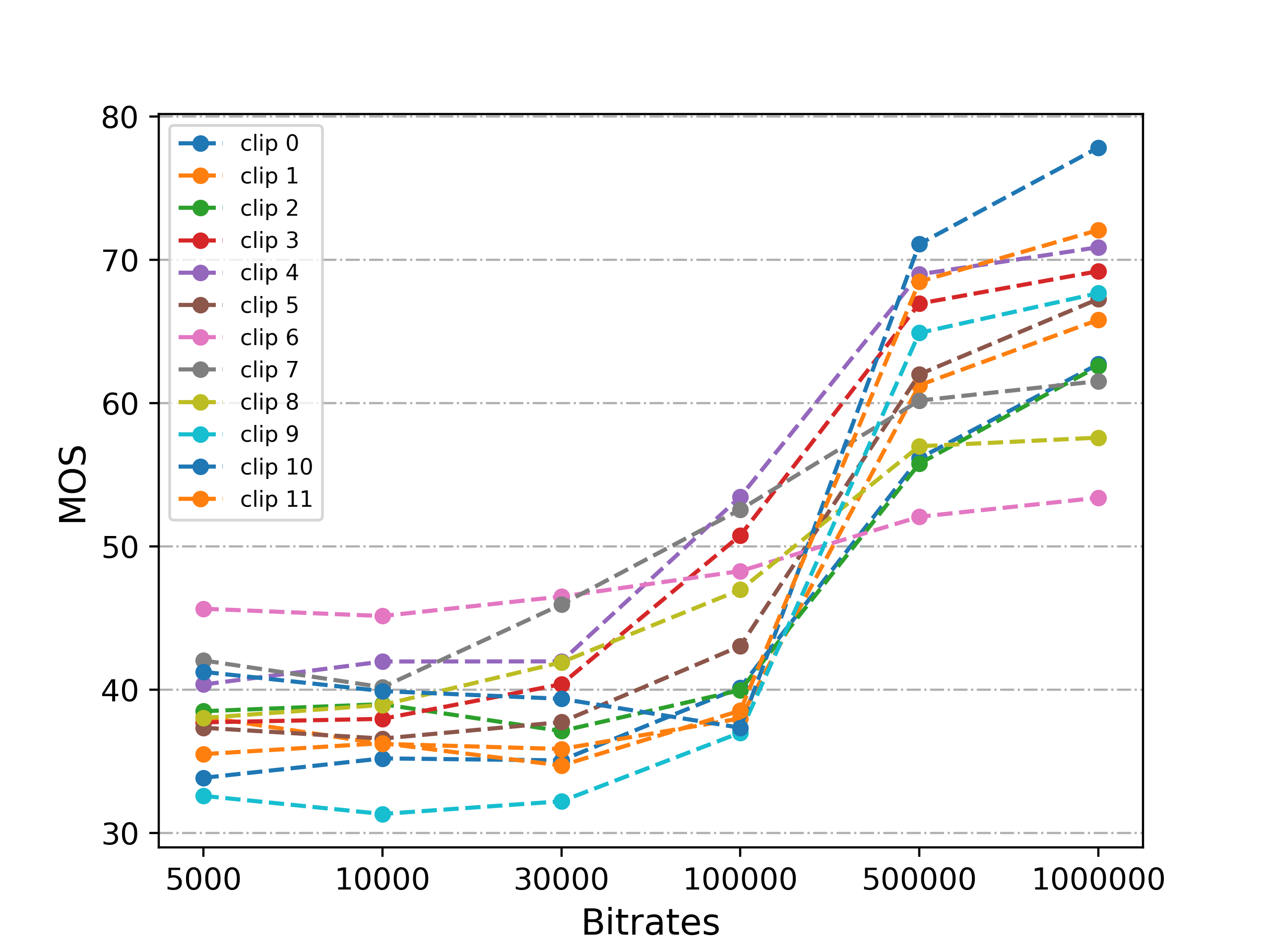}
    \caption{Predicted MOS of  720p videos}
    \label{fig:720p}
  \end{subfigure}

  \begin{subfigure}{0.42\linewidth}
    \includegraphics[width=1\textwidth]{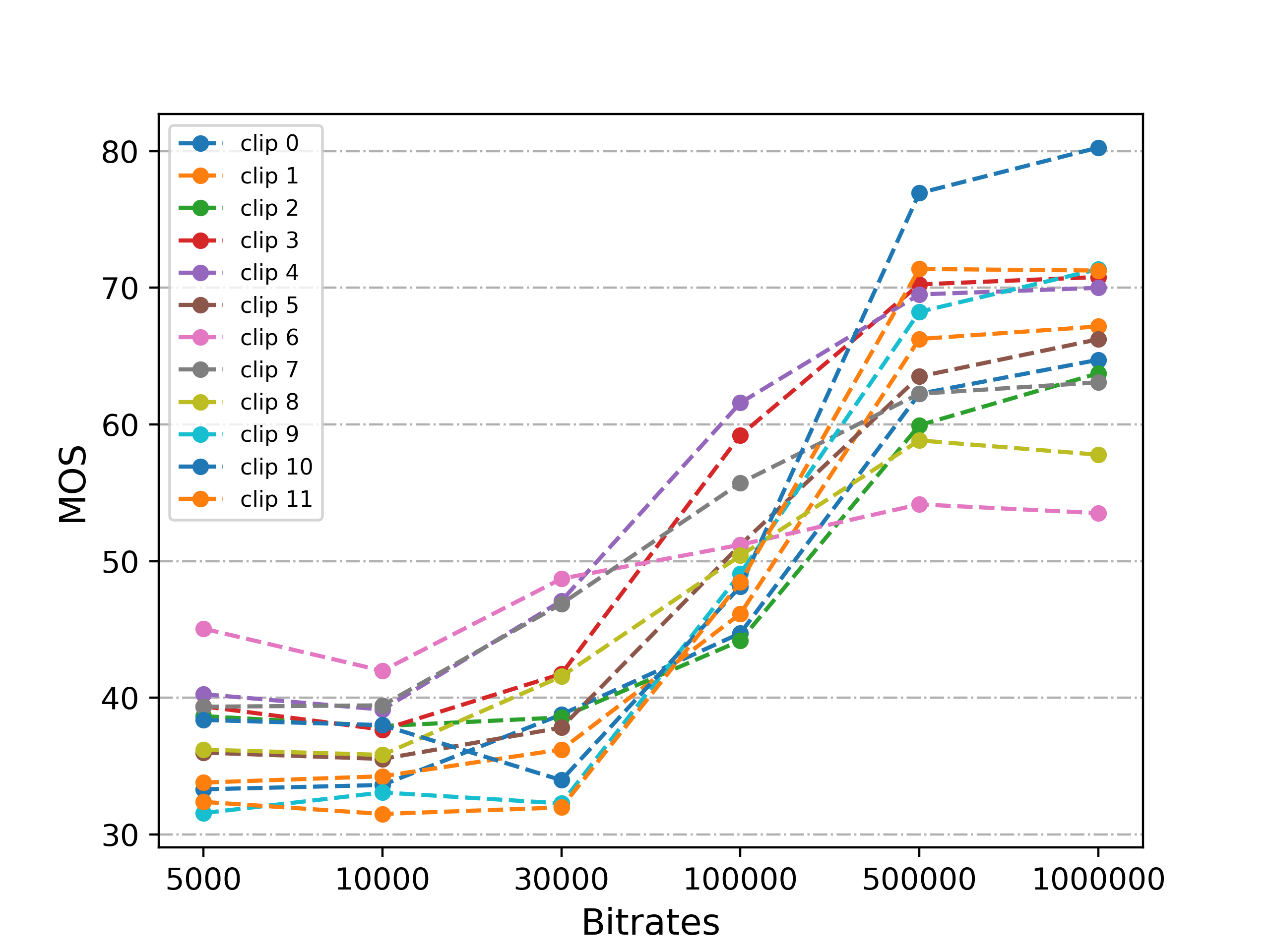}
    \caption{Predicted MOS of  480p videos}
    \label{fig:480p}
  \end{subfigure}
  \begin{subfigure}{0.42\linewidth}
    \includegraphics[width=1\textwidth]{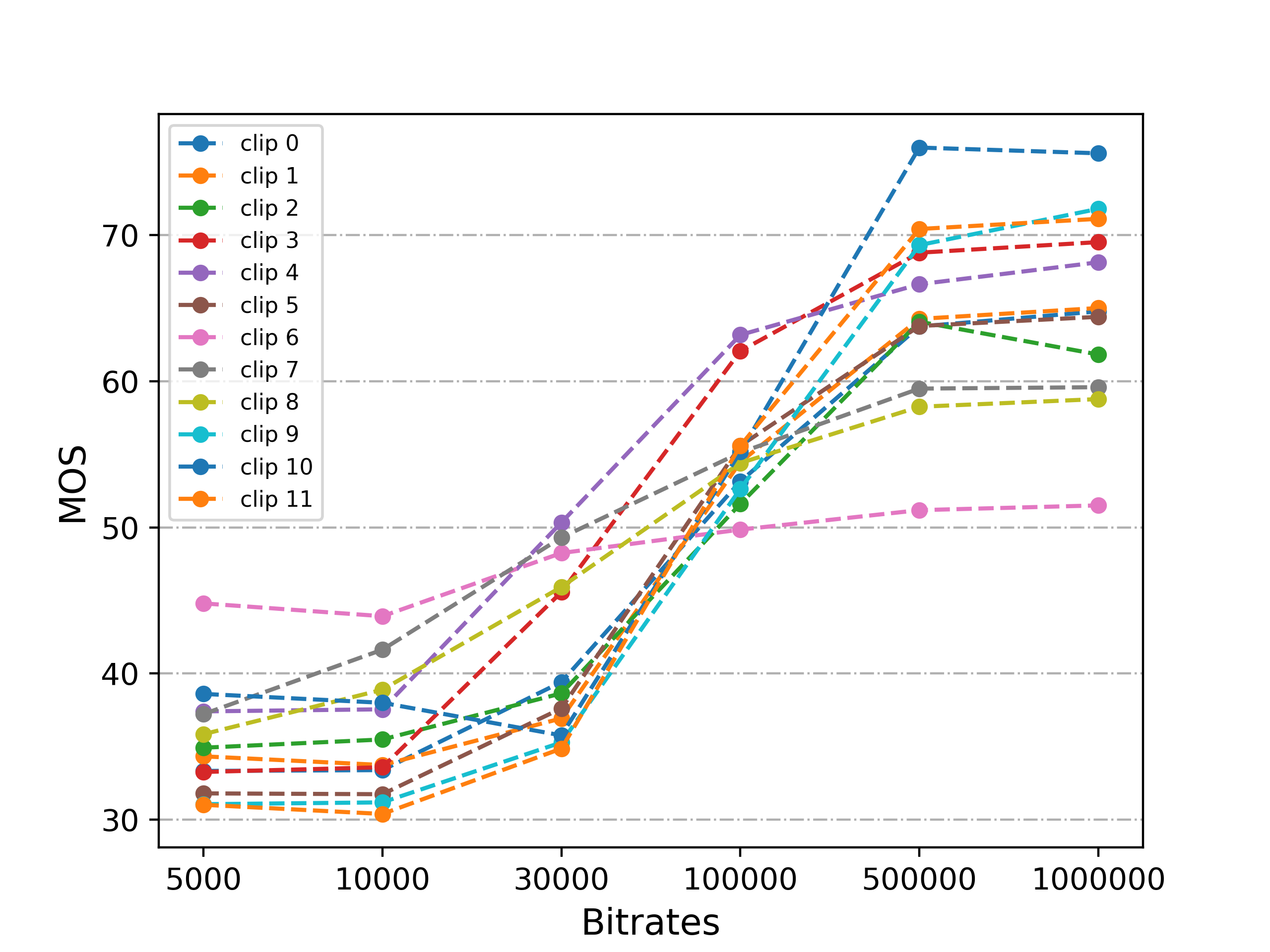}
    \caption{Predicted MOS of 360p videos}
    \label{fig:360p}
  \end{subfigure}
  \caption{The predicted Mean Opinion Score (MOS) consistently increases as the encoding bitrates increases for various resolutions, including 320p, 420p, 720p, and 1080p.}
  \label{fig:pgcvq}
\end{figure*}

\subsubsection{Heat-Map of videos}
Our dataset includes the Heat-Map information of movie trailers from YouTube. Heatmaps use a graph on the seek bar to highlight the most-watched parts of a video \cite{google}, which represents the human interests. In general, a heatmap divides a video into segments, each typically lasting 1 or 2 seconds, and assigns a score to each segment. The YouTube heatmap facilitates the identification of the most visually engaging scenes within a video.

\subsection{Study}

Our objectives are twofold: firstly, to evaluate whether the predicted score obtained from SB-VQA is consistent with the experience that higher encoding bitrates lead to better perceptual quality for PGC videos; secondly, to observe the relationship between the predicted perceptual quality and the human interests. To achieve the first goal, we randomly sampled $1,200$ videos, each with a length of $8-10$ seconds, from PGCVQ. Then, we made predictions on the $1,200$ videos with our proposed SB-VQA trained on LSVQ. For the second goal, we utilized the information obtained from YouTube heatmap. We study the relationship between VQA algorithms and video content.

\subsubsection{Quality of Pictures}

We make predictions on videos from PGCVQ with SB-VQA. \cref{fig:pgcvq} shows the results of $12$ randomly selected clips. “Clip a” in the four figures represents the same content, and is encoded with different bitrates under each resolution. It can be observed that videos with higher encoding bitrates have higher predicted scores, which is consistent with the fact that videos with high bitrates have better perceptual quality. The similar predictions of a clip with different resolutions may result from the gap between UGC/PGC data and the diverse distributions between datasets. 

Based on these results, we conclude that existing VQA algorithms can be useful for many PGC-based applications, such as VSR, denoising, old film restorations, which can benefit video enhancement related research.

\begin{figure*}[t]
  \centering
   \includegraphics[width=1\linewidth]{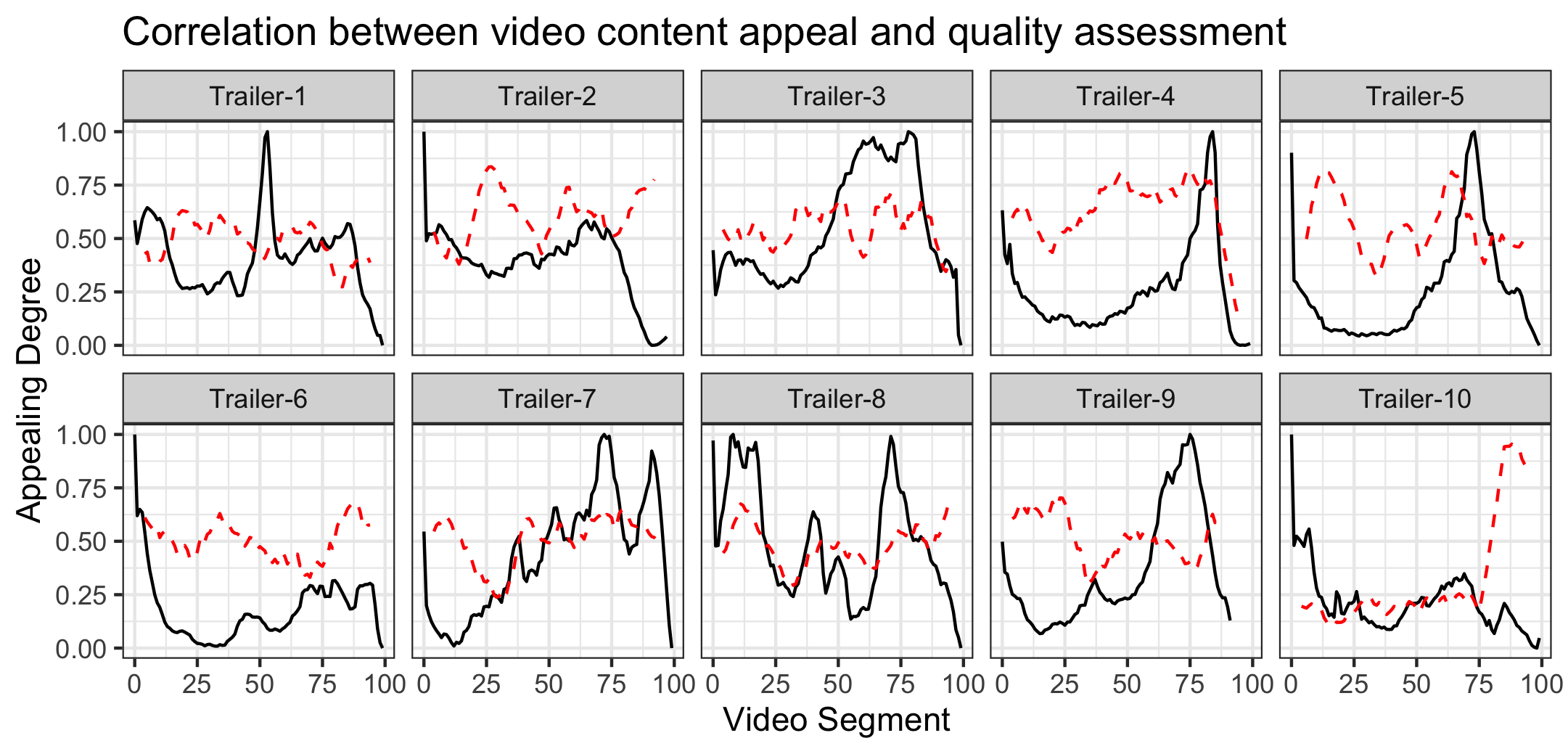}

   \caption{Correlation between YouTube heatmap-based content appeal (black line) and our predicted quality score (red dashed line) of example videos, suggesting strong forecasting capability of appealing scenes.}
   \label{fig:heatmap}
\end{figure*}

\subsubsection{Video Content}

We offer a novel perspective on the VQA task by considering the relationship between users' content preference and perceived video quality. To explore this relationship, we incorporate YouTube heatmaps of trailers into our experiments.

To observe trends, we scale our predicted quality score from $0$ to $1$ and compare it with the YouTube heatmaps for all video segments. Figure \ref{fig:heatmap} displays the trends of both YouTube heatmaps and our predicted quality scores for $10$ randomly selected movie trailers, where the y-axis represents the appealing degree of the video segments.

Based on our results, we observe that the YouTube heatmaps and predicted quality scores show good consistency, especially for Trailer 3, Trailer 4, Trailer 7, and Trailer 8. Upon examining the segments more closely, we find that action scenes and spectacular special effects tend to attract and engage viewers, resulting in higher peaks and higher predicted quality scores overall.

Interestingly, we observe opposite trends in our results for Trailer 2, Trailer 6, Trailer 9, and Trailer 10. Upon closer examination of these segments, we discover that some segments with high heatmap-based content appeal may be attributed to factors other than visual attractiveness, such as plot twists or unexpected developments in the story.

Taken together, our results suggest that existing VQA algorithms have the potential to analyze the richness of a video's content. While these algorithms primarily focus on the quality of the visuals, they also take into account the overall appeal of the video's content. We speculate that a multi-modal model incorporating semantic understanding could further improve VQA tasks.

\section{Conclusion}

To perform VQA tasks on video datasets with diverse enhancements, we propose a stack-based framework that can reduce bias and achieve better performance than the existing state-of-the-art. We further discuss applying VQA on PGC videos by building a movie trailer dataset. The results of our experiments show that VQA on PGC videos is feasible, and is beneficial to related research, e.g., VSR or restoration of classic films. Moreover, our analysis shows that VQA relies not only on quality of pictures but also on content. Semantic related research should be beneficial for VQA tasks. 

{\small
\bibliographystyle{ieee_fullname}

\bibliography{egbib}

\begin{thebibliography}{10}\itemsep=-1pt

\bibitem{google}
Youtube video player updates: Most replayed, video chapters, single loop \&amp;
  more - youtube community.

\bibitem{Avatar2022}
Avatar.
\newblock Avatar: The way of water | official teaser trailer, May 2022.

\bibitem{berns2019v3c1}
Fabian Berns, Luca Rossetto, Klaus Schoeffmann, Christian Beecks, and George
  Awad.
\newblock V3c1 dataset: an evaluation of content characteristics.
\newblock In {\em Proceedings of the 2019 on International Conference on
  Multimedia Retrieval}, pages 334--338, 2019.

\bibitem{bosse2017deep}
Sebastian Bosse, Dominique Maniry, Klaus-Robert M{\"u}ller, Thomas Wiegand, and
  Wojciech Samek.
\newblock Deep neural networks for no-reference and full-reference image
  quality assessment.
\newblock {\em IEEE Transactions on image processing}, 27(1):206--219, 2017.

\bibitem{chan2022basicvsr++}
Kelvin~CK Chan, Shangchen Zhou, Xiangyu Xu, and Chen~Change Loy.
\newblock Basicvsr++: Improving video super-resolution with enhanced
  propagation and alignment.
\newblock In {\em Proceedings of the IEEE/CVF conference on computer vision and
  pattern recognition}, pages 5972--5981, 2022.

\bibitem{Chen2016XGBoostAS}
Tianqi Chen and Carlos Guestrin.
\newblock Xgboost: A scalable tree boosting system.
\newblock {\em Proceedings of the 22nd ACM SIGKDD International Conference on
  Knowledge Discovery and Data Mining}, 2016.

\bibitem{ciancio2010no}
Alexandre Ciancio, Eduardo~AB da Silva, Amir Said, Ramin Samadani, Pere
  Obrador, et~al.
\newblock No-reference blur assessment of digital pictures based on
  multifeature classifiers.
\newblock {\em IEEE Transactions on image processing}, 20(1):64--75, 2010.

\bibitem{deng2009imagenet}
Jia Deng, Wei Dong, Richard Socher, Li-Jia Li, Kai Li, and Li Fei-Fei.
\newblock Imagenet: A large-scale hierarchical image database.
\newblock In {\em 2009 IEEE conference on computer vision and pattern
  recognition}, pages 248--255. Ieee, 2009.

\bibitem{dosovitskiy2020image}
Alexey Dosovitskiy, Lucas Beyer, Alexander Kolesnikov, Dirk Weissenborn,
  Xiaohua Zhai, Thomas Unterthiner, Mostafa Dehghani, Matthias Minderer, Georg
  Heigold, Sylvain Gelly, et~al.
\newblock An image is worth 16x16 words: Transformers for image recognition at
  scale.
\newblock {\em arXiv preprint arXiv:2010.11929}, 2020.

\bibitem{fang2020perceptual}
Yuming Fang, Hanwei Zhu, Yan Zeng, Kede Ma, and Zhou Wang.
\newblock Perceptual quality assessment of smartphone photography.
\newblock In {\em Proceedings of the IEEE/CVF Conference on Computer Vision and
  Pattern Recognition}, pages 3677--3686, 2020.

\bibitem{feichtenhofer2019slowfast}
Christoph Feichtenhofer, Haoqi Fan, Jitendra Malik, and Kaiming He.
\newblock Slowfast networks for video recognition.
\newblock In {\em Proceedings of the IEEE/CVF international conference on
  computer vision}, pages 6202--6211, 2019.

\bibitem{gao2023vdpve}
Yixuan Gao, Yuqin Cao, Tengchuan Kou, Wei Sun, Yunlong Dong, Xiaohong Liu,
  Xiongkuo Min, and Guangtao Zhai.
\newblock Vdpve: Vqa dataset for perceptual video enhancement.
\newblock {\em arXiv preprint arXiv:2303.09290}, 2023.

\bibitem{ghadiyaram2015massive}
Deepti Ghadiyaram and Alan~C Bovik.
\newblock Massive online crowdsourced study of subjective and objective picture
  quality.
\newblock {\em IEEE Transactions on Image Processing}, 25(1):372--387, 2015.

\bibitem{ghadiyaram2017capture}
Deepti Ghadiyaram, Janice Pan, Alan~C Bovik, Anush~Krishna Moorthy, Prasanjit
  Panda, and Kai-Chieh Yang.
\newblock In-capture mobile video distortions: A study of subjective behavior
  and objective algorithms.
\newblock {\em IEEE Transactions on Circuits and Systems for Video Technology},
  28(9):2061--2077, 2017.

\bibitem{he2016deep}
Kaiming He, Xiangyu Zhang, Shaoqing Ren, and Jian Sun.
\newblock Deep residual learning for image recognition.
\newblock In {\em Proceedings of the IEEE conference on computer vision and
  pattern recognition}, pages 770--778, 2016.

\bibitem{hosu2017konstanz}
Vlad Hosu, Franz Hahn, Mohsen Jenadeleh, Hanhe Lin, Hui Men, Tam{\'a}s
  Szir{\'a}nyi, Shujun Li, and Dietmar Saupe.
\newblock The konstanz natural video database (konvid-1k).
\newblock In {\em 2017 Ninth international conference on quality of multimedia
  experience (QoMEX)}, pages 1--6. IEEE, 2017.

\bibitem{hosu2020koniq}
Vlad Hosu, Hanhe Lin, Tamas Sziranyi, and Dietmar Saupe.
\newblock Koniq-10k: An ecologically valid database for deep learning of blind
  image quality assessment.
\newblock {\em IEEE Transactions on Image Processing}, 29:4041--4056, 2020.

\bibitem{kay2017kinetics}
Will Kay, Joao Carreira, Karen Simonyan, Brian Zhang, Chloe Hillier, Sudheendra
  Vijayanarasimhan, Fabio Viola, Tim Green, Trevor Back, Paul Natsev, et~al.
\newblock The kinetics human action video dataset.
\newblock {\em arXiv preprint arXiv:1705.06950}, 2017.

\bibitem{Korhonen2019TwoLevelAF}
Jari Korhonen.
\newblock Two-level approach for no-reference consumer video quality
  assessment.
\newblock {\em IEEE Transactions on Image Processing}, 28:5923--5938, 2019.

\bibitem{LeBlanc1996CombiningEI}
Michael LeBlanc and Robert Tibshirani.
\newblock Combining estimates in regression and classification.
\newblock {\em Journal of the American Statistical Association}, 91:1641--1650,
  1996.

\bibitem{li2022blindly}
Bowen Li, Weixia Zhang, Meng Tian, Guangtao Zhai, and Xianpei Wang.
\newblock Blindly assess quality of in-the-wild videos via quality-aware
  pre-training and motion perception.
\newblock {\em IEEE Transactions on Circuits and Systems for Video Technology},
  32(9):5944--5958, 2022.

\bibitem{Li2019Quality}
Dingquan Li, Tingting Jiang, and Ming Jiang.
\newblock Quality assessment of in-the-wild videos.
\newblock In {\em Proceedings of the 27th ACM International Conference on
  Multimedia}, pages 2351--2359, 2019.

\bibitem{li2021unified}
Dingquan Li, Tingting Jiang, and Ming Jiang.
\newblock Unified quality assessment of in-the-wild videos with mixed datasets
  training.
\newblock {\em International Journal of Computer Vision}, 129:1238--1257, 2021.

\bibitem{liao2022exploring}
Liang Liao, Kangmin Xu, Haoning Wu, Chaofeng Chen, Wenxiu Sun, Qiong Yan, and
  Weisi Lin.
\newblock Exploring the effectiveness of video perceptual representation in
  blind video quality assessment.
\newblock In {\em Proceedings of the 30th ACM International Conference on
  Multimedia}, pages 837--846, 2022.

\bibitem{liu2013bundled}
Shuaicheng Liu, Lu Yuan, Ping Tan, and Jian Sun.
\newblock Bundled camera paths for video stabilization.
\newblock {\em ACM transactions on graphics (TOG)}, 32(4):1--10, 2013.

\bibitem{liu2021swin}
Ze Liu, Yutong Lin, Yue Cao, Han Hu, Yixuan Wei, Zheng Zhang, Stephen Lin, and
  Baining Guo.
\newblock Swin transformer: Hierarchical vision transformer using shifted
  windows.
\newblock In {\em Proceedings of the IEEE/CVF international conference on
  computer vision}, pages 10012--10022, 2021.

\bibitem{liu2022video}
Ze Liu, Jia Ning, Yue Cao, Yixuan Wei, Zheng Zhang, Stephen Lin, and Han Hu.
\newblock Video swin transformer.
\newblock In {\em Proceedings of the IEEE/CVF conference on computer vision and
  pattern recognition}, pages 3202--3211, 2022.

\bibitem{mackin2018study}
Alex Mackin, Fan Zhang, and David~R Bull.
\newblock A study of high frame rate video formats.
\newblock {\em IEEE Transactions on Multimedia}, 21(6):1499--1512, 2018.

\bibitem{madhusudana2021subjective}
Pavan~C Madhusudana, Xiangxu Yu, Neil Birkbeck, Yilin Wang, Balu Adsumilli, and
  Alan~C Bovik.
\newblock Subjective and objective quality assessment of high frame rate
  videos.
\newblock {\em IEEE Access}, 9:108069--108082, 2021.

\bibitem{mittal2012no}
Anish Mittal, Anush~Krishna Moorthy, and Alan~Conrad Bovik.
\newblock No-reference image quality assessment in the spatial domain.
\newblock {\em IEEE Transactions on image processing}, 21(12):4695--4708, 2012.

\bibitem{mittal2012making}
Anish Mittal, Rajiv Soundararajan, and Alan~C Bovik.
\newblock Making a “completely blind” image quality analyzer.
\newblock {\em IEEE Signal processing letters}, 20(3):209--212, 2012.

\bibitem{nuutinen2016cvd2014}
Mikko Nuutinen, Toni Virtanen, Mikko Vaahteranoksa, Tero Vuori, Pirkko
  Oittinen, and Jukka H{\"a}kkinen.
\newblock Cvd2014—a database for evaluating no-reference video quality
  assessment algorithms.
\newblock {\em IEEE Transactions on Image Processing}, 25(7):3073--3086, 2016.

\bibitem{rippel2019learned}
Oren Rippel, Sanjay Nair, Carissa Lew, Steve Branson, Alexander~G Anderson, and
  Lubomir Bourdev.
\newblock Learned video compression.
\newblock In {\em Proceedings of the IEEE/CVF International Conference on
  Computer Vision}, pages 3454--3463, 2019.

\bibitem{Saad2014BlindPO}
Michele~A. Saad, Alan~Conrad Bovik, and Christophe Charrier.
\newblock Blind prediction of natural video quality.
\newblock {\em IEEE Transactions on Image Processing}, 23:1352--1365, 2014.

\bibitem{sinno2018large}
Zeina Sinno and Alan~Conrad Bovik.
\newblock Large-scale study of perceptual video quality.
\newblock {\em IEEE Transactions on Image Processing}, 28(2):612--627, 2018.

\bibitem{sun2022deep}
Wei Sun, Xiongkuo Min, Wei Lu, and Guangtao Zhai.
\newblock A deep learning based no-reference quality assessment model for ugc
  videos.
\newblock In {\em Proceedings of the 30th ACM International Conference on
  Multimedia}, pages 856--865, 2022.

\bibitem{tu2020comparative}
Zhengzhong Tu, Chia-Ju Chen, Li-Heng Chen, Neil Birkbeck, Balu Adsumilli, and
  Alan~C Bovik.
\newblock A comparative evaluation of temporal pooling methods for blind video
  quality assessment.
\newblock In {\em 2020 IEEE International Conference on Image Processing
  (ICIP)}, pages 141--145. IEEE, 2020.

\bibitem{Tu2020UGCVQABB}
Zhengzhong Tu, Yilin Wang, Neil Birkbeck, Balu Adsumilli, and Alan~Conrad
  Bovik.
\newblock Ugc-vqa: Benchmarking blind video quality assessment for user
  generated content.
\newblock {\em IEEE Transactions on Image Processing}, 30:4449--4464, 2020.

\bibitem{Tu2021RAPIQUERA}
Zhengzhong Tu, Xiangxu Yu, Yilin Wang, Neil Birkbeck, Balu Adsumilli, and
  Alan~Conrad Bovik.
\newblock Rapique: Rapid and accurate video quality prediction of user
  generated content.
\newblock {\em IEEE Open Journal of Signal Processing}, 2:425--440, 2021.

\bibitem{wang2019youtube}
Yilin Wang, Sasi Inguva, and Balu Adsumilli.
\newblock Youtube ugc dataset for video compression research.
\newblock In {\em 2019 IEEE 21st International Workshop on Multimedia Signal
  Processing (MMSP)}, pages 1--5. IEEE, 2019.

\bibitem{wang2021rich}
Yilin Wang, Junjie Ke, Hossein Talebi, Joong~Gon Yim, Neil Birkbeck, Balu
  Adsumilli, Peyman Milanfar, and Feng Yang.
\newblock Rich features for perceptual quality assessment of ugc videos.
\newblock In {\em Proceedings of the IEEE/CVF Conference on Computer Vision and
  Pattern Recognition}, pages 13435--13444, 2021.

\bibitem{wang2004image}
Zhou Wang, Alan~C Bovik, Hamid~R Sheikh, and Eero~P Simoncelli.
\newblock Image quality assessment: from error visibility to structural
  similarity.
\newblock {\em IEEE transactions on image processing}, 13(4):600--612, 2004.

\bibitem{wu2022fast}
Haoning Wu, Chaofeng Chen, Jingwen Hou, Liang Liao, Annan Wang, Wenxiu Sun,
  Qiong Yan, and Weisi Lin.
\newblock Fast-vqa: Efficient end-to-end video quality assessment with fragment
  sampling.
\newblock In {\em Computer Vision--ECCV 2022: 17th European Conference, Tel
  Aviv, Israel, October 23--27, 2022, Proceedings, Part VI}, pages 538--554.
  Springer, 2022.

\bibitem{Wu2022NeighbourhoodRS}
Haoning Wu, Chaofeng Chen, Liang Liao, Jingwen Hou, Wenxiu Sun, Qiong Yan,
  Jinwei Gu, and Weisi Lin.
\newblock Neighbourhood representative sampling for efficient end-to-end video
  quality assessment.
\newblock {\em ArXiv}, abs/2210.05357, 2022.

\bibitem{wu2022disentangling}
Haoning Wu, Liang Liao, Chaofeng Chen, Jingwen Hou, Annan Wang, Wenxiu Sun,
  Qiong Yan, and Weisi Lin.
\newblock Disentangling aesthetic and technical effects for video quality
  assessment of user generated content.
\newblock {\em arXiv preprint arXiv:2211.04894}, 2022.

\bibitem{yang2022maniqa}
Sidi Yang, Tianhe Wu, Shuwei Shi, Shanshan Lao, Yuan Gong, Mingdeng Cao, Jiahao
  Wang, and Yujiu Yang.
\newblock Maniqa: Multi-dimension attention network for no-reference image
  quality assessment.
\newblock In {\em Proceedings of the IEEE/CVF Conference on Computer Vision and
  Pattern Recognition}, pages 1191--1200, 2022.

\bibitem{Ying2020PatchVQU}
Zhenqiang Ying, Maniratnam Mandal, Deepti Ghadiyaram, Alan Bovik~University
  of~Texas~at Austin, and AI Facebook.
\newblock Patch-vq: ‘patching up’ the video quality problem.
\newblock {\em 2021 IEEE/CVF Conference on Computer Vision and Pattern
  Recognition (CVPR)}, pages 14014--14024, 2020.

\bibitem{ying2020patches}
Zhenqiang Ying, Haoran Niu, Praful Gupta, Dhruv Mahajan, Deepti Ghadiyaram, and
  Alan Bovik.
\newblock From patches to pictures (paq-2-piq): Mapping the perceptual space of
  picture quality.
\newblock In {\em Proceedings of the IEEE/CVF Conference on Computer Vision and
  Pattern Recognition}, pages 3575--3585, 2020.

\bibitem{zhai2008cross}
Guangtao Zhai, Jianfei Cai, Weisi Lin, Xiaokang Yang, Wenjun Zhang, and Minoru
  Etoh.
\newblock Cross-dimensional perceptual quality assessment for low bit-rate
  videos.
\newblock {\em IEEE Transactions on Multimedia}, 10(7):1316--1324, 2008.

\bibitem{zhang2018blind}
Weixia Zhang, Kede Ma, Jia Yan, Dexiang Deng, and Zhou Wang.
\newblock Blind image quality assessment using a deep bilinear convolutional
  neural network.
\newblock {\em IEEE Transactions on Circuits and Systems for Video Technology},
  30(1):36--47, 2018.

\end{thebibliography}
}

\end{document}